\documentclass{article}

\usepackage[nonatbib,preprint]{neurips}
\usepackage{algorithm}
\usepackage[utf8]{inputenc} % allow utf-8 input
\usepackage[T1]{fontenc}    % use 8-bit T1 fonts
\usepackage{hyperref}       % hyperlinks
\usepackage{url}            % simple URL typesetting
\usepackage{booktabs}       % professional-quality tables
\usepackage{amsfonts}       % blackboard math symbols
\usepackage{nicefrac}       % compact symbols for 1/2, etc.
\usepackage{microtype}      % microtypography
\usepackage{xcolor}         % colors
\usepackage{algorithmic}

\usepackage{amsmath}
\usepackage{amssymb}
\usepackage{mathtools}
\usepackage{amsthm}
\usepackage{amsmath}
\usepackage{algorithmic} 
\theoremstyle{plain}

\theoremstyle{definition}

\theoremstyle{remark}

\usepackage{graphicx}
\usepackage{subfigure}
\usepackage{listings}
\usepackage{tikz} % For drawing shapes

\definecolor{customgreen}{RGB}{119, 172, 48}
\newcommand{\greensquare}{\textcolor{customgreen}{\rule{1.5ex}{1.5ex}}}
\definecolor{customorange}{RGB}{217, 83, 25}
\newcommand{\orangesquare}{\textcolor{customorange}{\rule{1.5ex}{1.5ex}}}

\definecolor{customblue}{RGB}{77, 190, 238}

\newcommand{\bluesquare}{\textcolor{customblue}{\rule{1.5ex}{1.5ex}}}

\lstdefinestyle{mystyle}{
    backgroundcolor=\color{codebg},   
    commentstyle=\color{codegreen},
    keywordstyle=\color{codered},
    numberstyle=\tiny\color{codegray},
    stringstyle=\color{codepurple},
    basicstyle=\footnotesize,
    breakatwhitespace=false,         
    breaklines=true,                 
    captionpos=b,                    
    keepspaces=true,                 
    numbers=left,                    
    numbersep=5pt,                  
    showspaces=false,                
    showstringspaces=false,
    showtabs=false,                  
    tabsize=2
}
\definecolor{codebg}{rgb}{0.95, 0.95, 0.95} % Light grey background
\definecolor{codegreen}{rgb}{0,0.6,0}
\definecolor{codegray}{rgb}{0.5,0.5,0.5}
\definecolor{codepurple}{rgb}{0.58,0,0.82}
\definecolor{codered}{rgb}{0.6, 0.1, 0.1} % Red brick color
\usepackage{tikz} % For drawing shapes
\title{Probabilistic~Contrastive~Learning with Explicit Concentration on~the~Hypersphere}
\author{%
  Hongwei Bran Li \\
Athinoula A. Martinos Center \\
  MGH, HMS, Harvard University \\
  \texttt{holi2@mgh.harvard.edu} \\
  \And
  Cheng Ouyang\\
  Imperial College London \\
  \texttt{c.ouyang@imperial.ac.uk} \\
  \And
   Tamaz Amiranashvili \\
 Technical University of Munich \\
\texttt{tamaz.amiranashvili@uzh.ch} \\
  \And
  Matthew S. Rosen \\
Athinoula A. Martinos Center \\
  MGH, HMS, Harvard University \\
  \texttt{msrosen@mgh.harvard.edu} \\
    \And
    Bjoern Menze \\
  University of Zurich \\
  \texttt{bjoern.menze@uzh.ch} \\
  \And
  Juan Eugenio Iglesias \\
Athinoula A. Martinos Center \\
  MGH, HMS, Harvard University \\
  \texttt{jei@mit.edu} \\
}
\begin{document}
\maketitle
\begin{abstract}
Self-supervised contrastive learning has predominantly adopted deterministic methods, which are not suited for environments characterized by uncertainty and noise. 
This paper introduces a new perspective on incorporating uncertainty into contrastive learning by embedding representations within a spherical space, inspired by the von Mises-Fisher distribution (\(vMF\)). We introduce an unnormalized form of \(vMF\) and leverage the concentration parameter, \(\kappa \), as a direct, interpretable measure to quantify uncertainty \textit{explicitly}. This approach not only provides a probabilistic interpretation of the embedding space but also offers a method to calibrate model confidence against varying levels of data corruption and characteristics.
Our empirical results demonstrate that the estimated concentration parameter correlates strongly with the degree of unforeseen data corruption encountered at test time, enables failure analysis, and enhances existing out-of-distribution detection methods.
\end{abstract}

\section{Introduction}
\vspace{-0.15cm}
Self-supervised contrastive learning has significantly narrowed the gap between unsupervised and supervised learning across various domains, including vision~\cite{chen2020simple,chen2021exploring,caron2021emerging,zbontar2021barlow} and multimodal learning~\cite{hager2023best}. Despite these notable achievements, current methods still fall short in critical aspects necessary for decision-making in high-stakes applications. In domains such as medical diagnosis~\cite{azizi2021big} and autonomous driving~\cite{kaya2022uncertainty}, where decisions can have serious consequences, accurately estimating uncertainty, especially aleatoric uncertainty inherent to data, is essential.

Traditional contrastive learning methods are predominantly \emph{deterministic} and lack mechanisms to gauge uncertainty, limiting their utility in scenarios where understanding the model's confidence is crucial. Previous attempts to incorporate uncertainty estimation have primarily utilized Gaussian distributions~\cite{kingma2015variational,gal2016uncertainty,upadhyay2023probvlm}, which may not align well with hyperspherical contrastive representations~\cite{bachman2019learning,tian2020contrastive,he2020momentum,chen2021exploring}. Recent research has begun exploring geometric properties of contrastive representations~\cite{wang2020understanding,wang2021understanding,ge2023hyperbolic}, prompting a shift towards probabilistic models better suited to these spaces.

% Traditional contrastive learning methods are predominantly \emph{deterministic} and lack mechanisms to gauge uncertainty, limiting their utility in scenarios where understanding the model's confidence in its predictions is vital. Previous attempts to incorporate uncertainty estimation have primarily utilized Gaussian distributions~\cite{kingma2015variational,gal2016uncertainty,upadhyay2023probvlm}. However, Gaussian distributions may not be inherently aligned with the nature of contrastive learning representations, which are often situated in hyperspherical spaces. The advantages of learning contrastive representations on the unit hypersphere, including improved representation quality and interpretability, have led to their widespread adoption in mainstream contrastive learning methodologies~\cite{bachman2019learning,tian2020contrastive,he2020momentum,chen2021exploring}. Recognizing this, recent research has begun to explore the geometric properties of contrastive learning-based representations~\cite{wang2020understanding,wang2021understanding,ge2023hyperbolic}, prompting a shift towards alternative probabilistic models better suited to these geometric spaces for uncertainty estimation.

%%%%%%%%%
\begin{figure}[!t]
    \centering
    \includegraphics[width=0.80\textwidth]{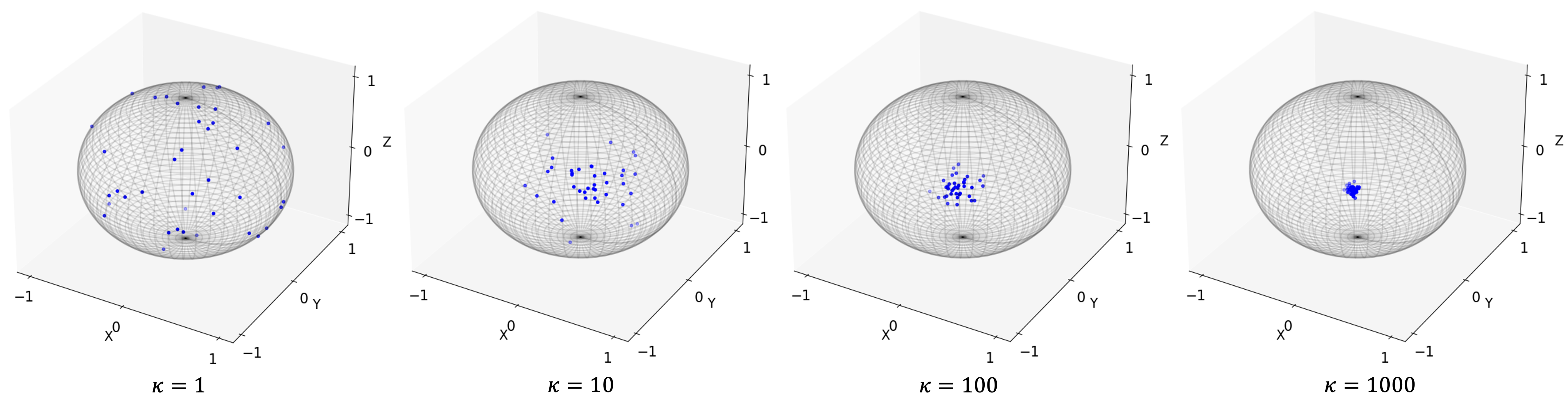}
    \vspace{-0.3cm}
    \caption{The von Mises-Fisher (\(vMF\)) distribution with a fixed mean vector and varied concentration values $\kappa$ on a sphere. Our framework equips contrastive learning representations to have the \(vMF\) distribution and utilizes the estimated $\kappa$'s as the indications of uncertainty.}
    \label{fig:kappa_visualization}
\end{figure}
%%%%%%%%%%%

Consequently, we turn our attention to the von Mises-Fisher (\(vMF\)) distribution~\cite{fisher1953dispersion}, a distribution adept at handling data on the sphere, thus aligning more closely with the intrinsic structure of most contrastive learning representations. The \(vMF\) distribution is parameterized by a mean direction~\(\boldsymbol{\mu}\) and a concentration parameter \(\kappa\). The $\kappa$ controls the spread of the distribution: when \(\kappa\) is high, the distribution is highly concentrated around \(\boldsymbol{\mu}\). As shown in Figure~\ref{fig:kappa_visualization}, this concentration means that the points sampled from the distribution are closely clustered around the mean, indicating lower dispersion. Therefore, \(\kappa\) can be viewed as a measure of uncertainty.

Recent advancements, such as the formulation proposed by \cite{kirchhof2023probabilistic}, have integrated Monte Carlo sampling with the contrastive loss, facilitating an \emph{implicit} learning of \(\kappa\). However, this approach encourages the \(\kappa\) values of paired positive samples to be similar, is not input-dependent, and does not capture \emph{explicit} uncertainty. In reality, aleatoric uncertainty varies across different instances \(x\) reflecting in distinct generative distributions \(p(y|x)\). 
% Hence, a relaxation on \(\kappa\) is essential, especially when images undergo strong and stochastic augmentations, where fine-grained uncertainty needs to be properly captured. 
In this work, we propose a novel probabilistic contrastive learning approach to model uncertainty by integrating the \(vMF\) distribution. Our probabilistic embedding alignment loss enforces \(\kappa\) as a purely \emph{input-dependent} parameter, allowing our method to estimate the inherent uncertainty of each data point, leading to more reliable uncertainty estimation.

In summary, \textbf{(1)}~we propose a \(vMF\)-inspired probabilistic contrastive learning framework that captures uncertainty in hyperspherical spaces. \textbf{(2)}~We develop an embedding alignment loss accounting for direction and concentration, compatible with existing contrastive learning methods while maintaining representation discriminativeness. \textbf{(3)}~We empirically demonstrate our framework's effectiveness in quantifying degrees of corruption and failure analysis during test time and potential in enhancing the representations for out-of-the-distribution (OOD) detection.

% In summary, \textbf{(1)}~we introduce a probabilistic contrastive learning framework inspired by the  \(vMF\) distribution, that effectively captures the uncertainty in hyperspherical embedding spaces and maintains the discriminativeness of representations. \textbf{(2)}~We develop a novel embedding alignment loss that accounts for both the direction and concentration of data representations on the hypersphere. This alignment loss is seamlessly compatible with existing contrastive learning approaches and the estimated concentration is used as the indication of uncertainty. \textbf{(3)}~Empirically, we demonstrate the effectiveness of our framework on robust learning. Moreover, we apply our approach to several mainstream contrastive learning methods, highlighting our method's potential and applicability in enhancing the reliability of contrastive learning representations.
\vspace{-0.15cm}
\section{Method}
\vspace{-0.15cm}
\subsection{Preliminaries on Contrastive Learning and the von Mises-Fisher Distribution}
\vspace{-0.15cm}
\textbf{Contrastive learning} learns to encode semantically similar data points close together and dissimilar points far apart in an embedding space in a \emph{deterministic} fashion. A common approach involves positive and negative pairs: given a data point \( \boldsymbol{x} \), two augmented views \( \boldsymbol{x}_i \) and \( \boldsymbol{x}_j \) are created. The objective is to maximize the similarity of these views (positive pairs) while minimizing the similarity with other data points (negative pairs). This can be formalized using a loss function like the widespread \emph{SimCLR} framework \cite{chen2020simple}. Its contrastive loss is formulated as: 
\begin{equation}
\label{eqn:equation_1}
\mathcal{L}_{\text{contrastive}} = -\log \frac{\exp(\text{sim}(\boldsymbol{z}_i, \boldsymbol{z}_j) / \tau)}{\sum_{k=1}^{N} \exp(\text{sim}(\boldsymbol{z}_i, \boldsymbol{z}_k) / \tau)},
\end{equation}
where \( \boldsymbol{z}_i \) and \( \boldsymbol{z}_j \) are the representations of \( \boldsymbol{x}_i \) and \( \boldsymbol{x}_j \) in embedding space, \( \text{sim}(\cdot) \) is a similarity measure (typically the \emph{cosine similarity}), \( \tau \) is a temperature scaling parameter, and the sum in the denominator is over one positive and \( N-1 \) negative pairs.

\textbf{The \(vMF\) distribution}~\cite{fisher1953dispersion} is a distribution on the unit sphere in \( \mathbb{R}^n \) to model data points on an $n$-dimensional hypersphere, making it suitable for spherical embedding spaces. Its probability density function is:
\begin{equation}
p(\boldsymbol{x}; \boldsymbol{\mu}, \kappa) = C(\kappa) \exp(\kappa \boldsymbol{\mu}^T \boldsymbol{x}),
\end{equation}
where \( \boldsymbol{x} \) is a point on the unit sphere, \( \boldsymbol{\mu} \) is the mean direction, \( \kappa \) is the scalar concentration parameter, and $C(\kappa) = \kappa^{n/2 - 1}  (2\pi)^{-n/2} / I_{n/2 - 1}(\kappa)$ is the normalization constant that ensures integration to one; \( I_{\nu} \) is the modified Bessel function~\cite{watson1922treatise} of the first kind at order \( \nu \). The concentration parameter \( \kappa \) determines the dispersion of the distribution around the mean direction \( \boldsymbol{\mu} \). A higher \( \kappa \) indicates less dispersion (more concentration around the mean), whereas a lower \( \kappa \) indicates more dispersion.

There is an \textbf{overflow issue} in \(vMF\) distribution, which arises due to the calculation of the normalization constant \(C(\kappa)\) \cite{banerjee2005clustering}. The modified Bessel function of the first kind \(I_{\nu}(\kappa)\) grows exponentially with increasing \(\kappa\), especially in high-dimensional spaces where the order \(\nu = \frac{n}{2} - 1\) can be large. This rapid growth can lead to numerical overflow, where the function values exceed the maximum representable value in floating-point arithmetic. Such an overflow issue can disrupt the stability and accuracy of computations, particularly during the training of models using gradient-based optimization methods. 

\subsection{Unnormalized \(vMF\) Distribution with a Regularization}
\label{sec:analysis_reg}
\textbf{An unnormalized and simplified form.} To circumvent the overflow issue discussed above, we consider an {unnormalized} form of the \(vMF\) distribution that omits the normalization constant and focuses on the core exponential term:
\begin{equation}
\psi(\boldsymbol{x}; \boldsymbol{\mu}, \kappa) = \exp(\kappa \boldsymbol{\mu}^\top \boldsymbol{x})
\end{equation}
This unnormalized form simplifies our model, making it computationally more efficient and stable, while still capturing the essential directional and concentration properties of the data. Although the unnormalized \(vMF\) is unbounded (for large kappa) and does not yield true probabilities, it is in practice an effective contrastive learning objective -- since it relies on relative comparisons rather than absolute values. The normalization constant $C(\kappa)$ for the \(vMF\) distribution inherently favors smaller values of $\kappa$ due to its mathematical properties. Specifically, the term in the log-likelihood acts as a natural regularizer on $\kappa$, encouraging the distribution to avoid extreme concentration and maintain a degree of uncertainty. However, removing $C(\kappa)$ necessitates an alternative technique to mimic this regularization effect. Below, we perform an analysis and propose a $\ell_2$ regularization on \(\kappa\).
%Empirically, we found that omitting it from the density function and introducing a regularization on $C(\kappa)$ maintains the ability to differentiate relative similarities between embeddings. %Further discussion on this issue can be found in Section \ref{sec:discussion_on_kappa} below.
%\(\{\boldsymbol{x}_i\}_{i=1}^N\)

\textbf{Analysis on a regularization on $\kappa$.} 
Consider the log-likelihood function of the \(vMF\)  distribution for a  data point \(\boldsymbol{x}\) on the unit sphere with mean direction \(\boldsymbol{\mu}\) and concentration parameter \(\kappa\):
\begin{equation}
\mathcal{L}(\boldsymbol{\mu}, \kappa) =  \log p(\boldsymbol{x}; \boldsymbol{\mu}, \kappa) =   \log C(\kappa) + \kappa \boldsymbol{\mu}^\top \boldsymbol{x} 
\end{equation}
% When \(C(\kappa)\) is included, it inherently discourages large \(\kappa\) values due to the properties of the Bessel function. Removing \(C(\kappa)\) necessitates an alternative form of regularization to avoid unbounded \(\kappa\) values, which can lead to numerical instability.
The normalization constant \(C(\kappa)\) involves the modified Bessel function \(I_{p/2-1}(\kappa)\), which grows exponentially with \(\kappa\). The term \(\log C(\kappa)\) in the log-likelihood acts as a natural penalty on \(\kappa\):
\begin{equation}
\log C(\kappa) = (p/2-1) \log \kappa - \log I_{p/2-1}(\kappa) - \frac{p}{2} \log (2\pi)
\end{equation}
For large \(\kappa\), \(I_{p/2-1}(\kappa) \approx \frac{e^\kappa}{\sqrt{2\pi \kappa}}\), thus:
\begin{equation}
\log C(\kappa) \approx (p/2-1) \log \kappa - \kappa - \log (\sqrt{2\pi \kappa})
\end{equation}
This approximation shows that \(\log C(\kappa)\) introduces a term \(-\kappa\), which effectively penalizes large \(\kappa\) values. The \(- \log (\sqrt{2\pi \kappa})\) term contributes a smaller additional penalty. By introducing $\ell_2$ regularization, we can mimic this penalization:
\begin{equation}
\mathcal{L}_{\text{reg}} = - \lambda \kappa^2
\end{equation}
Here, \(\lambda\) is a regularization parameter controlling the strength of the penalty. %This term helps constrain \(\kappa\), preventing it from becoming excessively large.
In high-dimensional spaces, large \(\kappa\) values can cause numerical instability due to the exponential terms in the \(vMF\) distribution. $\ell_2$ regularization on \(\kappa\) helps maintain stability by keeping \(\kappa\) values moderate, ensuring that gradients do not explode during optimization.

\textbf{Probabilistic interpretation.}
From a Bayesian perspective, regularization can be interpreted as placing a prior on the parameter. An $\ell_2$ regularization term corresponds to a Gaussian prior on \(\kappa\):
\begin{equation}
P(\kappa) \propto \exp(-\lambda \kappa^2)
\end{equation}
This prior assumes that \(\kappa\) takes smaller values, aligning with the nature of the  \(vMF\) distribution to avoid extreme concentrations.

% The cosine similarity, used in existing contrastive learning methods~\cite{chen2020simple,chen2021exploring,grill2020bootstrap}, measures the angle between two representations independent of their magnitudes, is thus amenable to probabilistic representations with \(vMF\) distribution.
% In high-dimensional spaces, the \(vMF\) distribution captures how data points are dispersed around this mean direction on the hypersphere, as illustrated in Figure~\ref{fig:kappa_visualization}. 
% Here, we adapt the \(vMF\) distribution to this setting to allow for a {probabilistic interpretation. %

\subsection{Probabilistic Contrastive Learning on the Hypersphere}

% \subsubsection{Hypersphere and Contrastive Learning}
\textbf{Unit sphere normalization.} First, we enhance the sensitivity of the contrastive learning representation to angular differences by projecting each data point \( \boldsymbol{x} \) onto the unit sphere, defined as:
$\boldsymbol{z} = f(\boldsymbol{x}) / \|f(\boldsymbol{x})\|$, 
where \( f(\boldsymbol{x}) \) is the representation obtained with an encoder network \( f(\cdot) \), and \( \|f(\boldsymbol{x})\| \) is its Euclidean norm. This mapping to the unit sphere ensures that all comparisons between embeddings are based solely on their directionality, which is inherently measured by cosine similarity in existing contrastive learning~\cite{chen2020simple,chen2021exploring,grill2020bootstrap}, aligning perfectly with our geometric constraints.

Then, we incorporate the \(vMF\) distribution into a contrastive learning framework that leverages this hyperspherical embedding. The \(vMF\) distribution is ideally suited for embeddings that are normalized in this manner and provide a probabilistic interpretation of the dispersion of points around a mean direction. % which is critical for understanding and optimizing the behavior of our contrastive losses within this constrained embedding space.

% \subsubsection{Probabilistic Contrastive Learning with Embedding Alignment} Inspired by the \(vMF\) distribution, we model the uncertainty in the learned representations using \(\kappa\).
\textbf{Probabilistic embedding alignment.} We leverage the directional consistency between data pairs and enforce \(\kappa\) to be a purely \emph{input-dependent} parameter. 

Let \(\boldsymbol{x}\) be an input image batch from a training set. Two augmented views of \(\boldsymbol{x}\), denoted as \(\boldsymbol{x}_1\) and \(\boldsymbol{x}_2\), are generated using a predefined augmentation pipeline. This pipeline typically includes transformations such as random cropping, resizing, color jittering, and horizontal flipping~\cite{chen2020simple}.

A neural network model \( f(\cdot) \) processes each augmented view and outputs an \(d\)-dimensional mean directional vector \( \boldsymbol{\mu'} \) and a scalar concentration parameter \( \kappa \). Let \( \boldsymbol{\mu} \) be the normalized vector: \(\boldsymbol{\mu} = \boldsymbol{\mu'} / \|\boldsymbol{\mu'}\|\). The embedding spaces are then represented as \( (\boldsymbol{\mu}_1, \kappa_{1}) = f(\boldsymbol{x}_1) \) and \( (\boldsymbol{\mu}_2, \kappa_{2}) = f(\boldsymbol{x}_2)\), where \(\kappa_1\) and \(\kappa_2\) are non-negative. 

In the contrastive learning context, we propose a \textit{probabilistic embedding alignment loss} for the distributions of two augmented views (positive pairs). The alignment of \( \boldsymbol{\mu}_2 \) given (\( \boldsymbol{\mu}_1, \kappa_{1}\)) and \( \boldsymbol{\mu}_1 \) given (\( \boldsymbol{\mu}_2, \kappa_{2}\)) can be formulated as:
\begin{equation}
\label{eqn:vmf}
    L_{\text{a}}( \boldsymbol{\mu}_1, \kappa_{1}, \boldsymbol{\mu}_2, \kappa_{2}) 
    = \exp[(\kappa_1 + \kappa_{2}) \cdot \boldsymbol{\mu}_1^T \boldsymbol{\mu}_2] \propto \exp(\kappa_{1} \cdot \cos(\theta)) \cdot \exp(\kappa_{2} \cdot \cos(\theta)),
\end{equation}
where \( \theta \) is the angle between \( \boldsymbol{\mu}_1 \) and \( \boldsymbol{\mu}_2 \), and \( \cos(\theta) \) can be computed as the dot product between \( \boldsymbol{\mu}_1 \) and \( \boldsymbol{\mu}_2 \) due to their normalization. The loss is then defined as the negative log-alignment:
\begin{equation}
\label{eqn:alignment}
    \mathcal{L}_{\text{align}} = - \lambda_{\text{align}} \cdot (\kappa_1 + \kappa_{2}) \boldsymbol{\mu}_1^T \boldsymbol{\mu}_2,
\end{equation}
where \( \lambda_{\text{align}} \) controls the strength of the loss. This loss emphasizes the exponential alignment of embeddings based on their dot product, scaled by the sum of their concentration parameters. Unlike the \emph{MC-InfoNCE} loss~\cite{kirchhof2023probabilistic}, our loss \emph{directly} links the strength of the alignment to the uncertainty of the embeddings, as represented by \(\kappa\). Intuitively, the alignment loss encourages the encoder to learn embeddings with a desired property: the mean directions \( \boldsymbol{\mu}_1 \) and \( \boldsymbol{\mu}_2 \) are closely aligned when the representations are certain (i.e., high \(\kappa\) values) while treating \(\kappa_1\) and \(\kappa_2\) as variables capturing different degrees of uncertainties. Hence, it relaxes the unrealistic assumption that \(\kappa\) values from positive pairs are expected to be \emph{similar} as in previous work~\cite{kirchhof2023probabilistic}. We provide an analysis on the gradient behavior of Eq.~\ref{eqn:alignment} in the Appendix.

Note that, due to the unnormalized \(vMF\) distribution, the loss in Eq.~\ref{eqn:alignment} can be infinitely small if \(\kappa_1\) and \(\kappa_2\) grow unboundedly. Therefore, as discussed in Sec. \ref{sec:analysis_reg}, we regularize \(\kappa\) by an $\ell_2$-norm regularization loss:
\begin{equation}
\label{eqn:l2_reg}
    \mathcal{L}_{\text{reg}}(\kappa_1, \kappa_2) = \lambda_{\text{reg}} \cdot (\kappa_{1}^2 + \kappa_{2}^2),
\end{equation}
where \( \lambda_{\kappa} \) is a hyperparameter that controls the strength of the regularization.

\textbf{The final loss.} To maintain the discriminativeness of the embeddings, we employ the original deterministic contrastive loss applied to the mean direction vectors from positive pairs and negative pairs, computed using Eq.~\ref{eqn:equation_1}. We combine the proposed \emph{embedding alignment} loss, the $\ell_2$-norm regularization, and the \emph{SimCLR} contrastive loss~\cite{chen2020simple}:
\begin{equation}
    \mathcal{L}_{\text{total}} = \mathcal{L}_{\text{align}}(\kappa_1, \kappa_2) + \mathcal{L}_{\text{reg}}(\kappa_1, \kappa_2) + \mathcal{L}_{\text{contrastive}}(\boldsymbol{x}_1, \boldsymbol{x}_2).
\end{equation}
This combined loss function enables the model to learn uncertainty-aware and discriminative embeddings. Notably, our proposed alignment loss requires only positive pairs, making it plug-and-play compatible with most contrastive learning methods. The optimization is detailed in Algorithm~\ref{alg:prob_contrastive_learning}. 
\begin{algorithm}[t]
\caption{Probabilistic Contrastive Learning with Embedding Alignment}
\label{alg:prob_contrastive_learning}
\begin{algorithmic}[1] % Numbering lines
\FOR{each batch $\{\boldsymbol{x}^i\}_{i=1}^{N}$}
 % \STATE 
    \STATE $\{(\boldsymbol{x}_1^i, \boldsymbol{x}_2^i)\}_{i=1}^{N} \leftarrow \text{Augment} (\{\boldsymbol{x}^i\}_{i=1}^{N})$ \textcolor{magenta}{\% Get positive pairs via augmentations}
    \FOR{each positive pair $(\boldsymbol{x}_1, \boldsymbol{x}_2)$ in the batch}
    % \STATE 
        \STATE $(\boldsymbol{\mu}_1, \kappa_{1}), (\boldsymbol{\mu}_2, \kappa_{2}) \leftarrow {f}(\boldsymbol{x}_1), {f}(\boldsymbol{x}_2)$ \textcolor{magenta}{\% Get mean directions and concentrations}
        % \STATE $(\boldsymbol{\mu}_2, \kappa_{2}) \leftarrow {f}(\boldsymbol{x}_2)$
    % \STATE 
        \STATE $\kappa_1, \kappa_2 \leftarrow \text{Softplus}(\kappa_{1}, \kappa_{2})$ \textcolor{magenta}{\%  Ensure positive $\kappa$ values}
        % \STATE $\kappa_2 \leftarrow \text{Softplus}(\kappa_{2})$
        \STATE Compute $\mathcal{L}_{\text{align}}$ with Eq. \ref{eqn:alignment}
        \STATE Compute $\mathcal{L}_{\text{contrastive}}$ using $(\boldsymbol{\mu}_1, \boldsymbol{\mu}_2)$
        \STATE $\mathcal{L}_{\text{reg}}(\kappa) \leftarrow \lambda_{\kappa} \cdot (\kappa_{1}^2 + \kappa_{2}^2)$
        \STATE $\mathcal{L}_{\text{total}} \leftarrow \mathcal{L}_{\text{align}} + \mathcal{L}_{\text{contrastive}} + \mathcal{L}_{\text{reg}}$
        \STATE Backpropagate and update model parameters
    \ENDFOR
\ENDFOR
\end{algorithmic}
\end{algorithm}

% \subsubsection{Discussion on $C(\kappa)$ and the $L^{2}$-norm}
% \label{sec:discussion_on_kappa}
% While $C(\kappa)$ ensures the \(vMF\) distribution integrates to one over the sphere's surface, it causes computational instability in high-dimensional calculations, particularly for large values of the concentration parameter $\kappa$. 

% However, the design of the alignment loss naturally encourages $\kappa$ to become unboundedly large. Our choice to employ $L^{2}$-norm regularization on $\kappa$ can be seen as an effort to implicitly mirror this aspect of the \(vMF\) distribution's behavior in a computationally stable and tractable manner. The regularization term directly penalizes large values of $\kappa$, effectively preventing the model from becoming overly confident and ensuring that the representations maintain a realistic level of dispersion on the hypersphere. This approach aligns with the natural inclination of the log-likelihood term to favor smaller $\kappa$ values, albeit in a more controlled fashion.

\section{Related Work}
\textbf{Representation learning on the unit hypersphere} has its advantages in representation quality and interpretability~\cite{nickel2017poincare, davidson2018hyperspherical}. Theoretical analysis has shown that such methods learn alignment and uniformity properties asymptotically on the hypersphere \cite{wang2020understanding}. It has been therefore widely adopted by the popular contrastive learning approaches \cite{bachman2019learning,tian2020contrastive,he2020momentum,chen2021exploring}. Hyperspherical latent spaces in variational autoencoders have demonstrated superior performance over Euclidean counterparts~\cite{davidson2018hyperspherical,xu2018spherical}.  
%
% Direct matching of uniformly sampled points on the unit hypersphere yielding robust representations~\cite{bojanowski2017unsupervised}. 
Hyperspherical face embeddings have outperformed their unnormalized counterparts~\cite{liu2017sphereface,wang2017normface}. Recently, contrastive learning on the hypersphere has been shown effective in out-of-distribution detection \cite{ming2022exploit}.
The consistent empirical success across diverse applications and nice geometric properties underscores the hypersphere's uniqueness as a feature space. In the context of our work, we extend this exploration to the realm of uncertainty estimation within these hyperspherical spaces.

\textbf{Aleatoric uncertainty} is inherent in many vision problems, such as object recognition \cite{kendall2017uncertainties,shi2019probabilistic} and semantic segmentation \cite{monteiro2020stochastic,kahl2024values}, where stochasticity in image acquisition (\textit{e.g.,} noise and imaging artifacts) incurs uncertainties in prediction. Other tasks with ambiguous input data include 3D reconstruction from 2D input \cite{chen2021monorun} or from noisy sensor \cite{meech2021algorithm}.
%In real-world scenarios, data corruption such as Gaussian noise introduces additional aleatoric uncertainty which can be reflected in its levels of corruption \cite{hendrycks2019benchmarking}. 
To facilitate the systematic study of aleatoric uncertainty, the widely-applied benchmark proposed by \cite{hendrycks2019benchmarking} quantifies the severity of data corruptions (\textit{e.g.,} imaging noise, distortions caused by compression, etc.) into different corruption levels \cite{hendrycks2019benchmarking}.
In this work, we demonstrate that the estimated concentration parameters $\kappa$'s closely \emph{correlate} with the corruption levels. %This key property is particularly useful for out-of-distribution detection \cite{vyas2018out,zhang2023openood}.  

\textbf{Probabilistic embedding} are emerging approaches that involve encoders generating distributions within the latent space, rather than deterministic point estimates. Such approaches to probabilistic embeddings diverge into two primary categories: The first method transforms traditional loss functions into probabilistic formats by aggregating the entire loss across the spectrum of predicted probabilistic embeddings \cite{scott2021mises,roads2021enriching,kirchhof2023probabilistic}. 
Another strategy employs distribution-to-distribution metrics to substitute the conventional point-to-point distances in loss calculations, with the Expected Likelihood Kernel \cite{shi2019probabilistic} standing out as a particularly effective technique. Notably, it has recently shown its efficacy even in contexts involving high-dimensional embedding spaces~\cite{kirchhof2022nonisotropic}. Recently, a Monte-Carlo sampling-based \emph{InfoNCE} loss \cite{kirchhof2023probabilistic} was proposed to train the encoder to predict probabilistic embeddings and to learn the correct posteriors. In our work, we present a fresh perspective on modeling such a probabilistic embedding by introducing the unnormalized \(vMF\) and a regularization term, resulting in an effective and computationally efficient framework. 

\section{Experiments and Results}
% In this section, we begin by detailing the experimental setup, evaluation metrics, comparative baselines, and the training regimen. Subsequently, we present empirical findings and discussions.

\subsection{Experimental Setup}
% \subsubsection{Tasks and Metrics}
\textbf{Quantifying the level of data corruption.} CIFAR-10-C~\cite{hendrycks2018benchmarking} is a well-established benchmark dataset for evaluating model robustness in a controlled environment. It contains 18 image corruption types based on the original CIFAR-10~\cite{krizhevsky2009learning}. Our key assumption is that the corrupted data have higher inherent aleatoric uncertainty compared to the uncorrupted one. We therefore assume that higher degrees of corruption would result in higher uncertainties (lower concentration $\kappa$). We use \emph{Spearman Correlation} as an evaluation metric to quantify if a model could capture this connection. We use the non-parametric, ranking-based Spearman correlation (rather than Pearson) as the relationship between the variables is highly nonlinear (i.e., we test for their monotonicity). Some corruptions are shown in Figure \ref{fig:kappa_levels}.

\textbf{OOD detection.}
From CIFAR-10, CIFAR-100, and MNIST \cite{lecun1998mnist}, we generate six in-domain and out-of-domain pairs for the OOD detection tasks, as shown in Table 2. Area Under the Receiver Operating Characteristic curve (AUROC) is used for the detection accuracy following the practice from \cite{kuan2022back}. For this task, we train three different models on the three domains from scratch. The learned $\kappa$ is treated as a one-dimensional feature (or anomaly score) to enhance the features for OOD detection.

% \vspace{-0.2cm}
\subsection{Baselines}
To quantify the uncertainty in representations, we compare our method with the following baselines. 

\textbf{Model ensembles.}
We train multiple models and then evaluate the empirical variance in their representations. The variance among these models' representations is used as a measure of uncertainty. Practically, each model in the ensemble can be trained with stochasticity (different initializations, etc.) on the same dataset or from different training epochs~\cite{huang2016snapshot}. High variance indicates less certainty in the representations.

Let \( f_{i,d}(x) \) denote the prediction of the \( i \)-th model in the ensemble for input \( x \) at the \( d \)-th feature. For an ensemble of \( M \) models and a feature dimension \( D \), the average uncertainty \( U_{\text{avg}}(x) \) for input \( x \) is given by:
$U_{\text{avg}}(x) = \frac{1}{D} \sum_{d=1}^{D} \left( \frac{1}{M} \sum_{i=1}^{M} \left( f_{i,d}(x) - \bar{f}_d(x) \right)^2 \right)$, 
where \( \bar{f}_d(x) \) is the average prediction for input \( x \) at the \( d \)-th feature: $\bar{f}_d(x) = \frac{1}{M} \sum_{i=1}^{M} f_{i,d}(x)$.

\noindent 
\textbf{Monte Carlo (MC) dropout \cite{gal2016dropout}} quantifies uncertainty by enabling dropout during inference and running multiple forward passes through the network which can be seen as sampling from an approximate posterior distribution of the model weights. The variance across these forward passes is used to estimate the uncertainty.

Let \( g_{j,d}(x) \) represent the prediction of the model on the \( j \)-th forward pass with dropout for input \( x \) at the \( d \)-th feature. For \( N \) forward passes and feature dimension \( D \), the average uncertainty \( V_{\text{avg}}(x) \) for input \( x \) is: $V_{\text{avg}}(x) = \frac{1}{D} \sum_{d=1}^{D} \left( \frac{1}{N} \sum_{j=1}^{N} \left( g_{j,d}(x) - \bar{g}_d(x) \right)^2 \right)$, 
where \( \bar{g}_d(x) \) is the average prediction across the \( N \) forward passes for input \( x \) at the \( d \)-th feature: $\bar{g}_d(x) = \frac{1}{N} \sum_{j=1}^{N} g_{j,d}(x)$.

\textbf{MCInfoNCE \cite{kirchhof2023probabilistic}} involves an MC sampling process to model the probabilistic nature of embeddings via a reparametrization
trick for $vMF$ distribution~\cite{davidson2018hyperspherical}. This sampling process allows for the approximation of expectations over the latent space, which is essential in scenarios where direct calculation of these expectations is intractable. Since our method works with \emph{SimCLR} contrastive loss, for a fair comparison, we adapt their \emph{InfoNCE}-based \cite{oord2018representation} loss to the \emph{SimCLR} contrastive loss and use the same architectures for the mean direction \(\boldsymbol{\mu}\) and \(\kappa\). However, we still term it as \emph{MC-InfoNCE} loss for literature consistency in the following experiments. The codes of the implementation are in Appendix~\ref{MC-InfoNCE}. 
%By employing MC sampling, the \emph{MC-InfoNCE} loss can effectively incorporate the probabilistic distribution of embeddings into the contrastive learning framework.

% \vspace{-0.2cm}
\subsection{Training}
% \vspace{-0.2cm}
\textbf{Architecture.} The encoder network contains two projection heads for the mean direction \(\boldsymbol{\mu}\) and \(\kappa\) based on \emph{ResNet50}~\cite{he2016deep}.  
% In our proposed architecture, the concentration parameter \(\kappa\) is parameterized by a layer head, integrated into a \emph{ResNet50} \cite{he2016deep}. 
The projection head for the \(\boldsymbol{\mu}\) is realized through a sequential arrangement of layers, starting with a linear transformation from the 2048-dimensional \emph{ResNet50} feature space to an intermediate 512-dimensional space, followed by batch normalization and \emph{ReLU} activation, and finally projecting down to a $d$-dimensional representation. $d$ is set to 128 for all experiments except the study on dimension in Table \label{tab:dim}. 
In parallel, the \(\kappa\) parameter is estimated through a separate head, mirroring the structure but diverging in its final output to produce a single scalar value per input. \(\kappa\) is then passed through a \emph{softplus} function \cite{NairHinton2010}, ensuring its non-negativity and adherence to the constraints of a concentration parameter in a probabilistic setting. The codes of the neural architecture are in Appendix \ref{architecture}.

\textbf{Optimization.} Following the \emph{SimCLR} configuration, our data augmentation includes random cropping, resizing, color jittering, and horizontal flipping. We train all models on CIFAR-10's training set for 1000 epochs to quantify corruption levels. Hyper-parameters $\lambda_\text{align}$ and $\lambda_\text{reg}$ are adjusted for optimal training loss and stability, with $\lambda_\text{reg}$ fixed at 0.005 across experiments due to observed training stability. The $\lambda_\text{align}$ parameter, dictating alignment loss strength, inversely affects representation discriminativeness and, if increased, may cause training instability. A practical approach involves starting with a low value, like 0.01, and incrementally adjusting up to a saturation point where the total training loss stabilizes; here, $\lambda_\text{align}$ is set to 0.05 for \emph{SimCLR}. %Training codes are available in Appendix \ref{training}.

\begin{table*}[t]
\caption{\textbf{Spearman correlation between kappa values and the levels of corruption}. As the severity of corruption increases, \( \kappa \) decreases, implying higher uncertainty in the representations. $+$ and $-$ indicate that the correlations are expected to be \emph{positive} and \emph{negative}, respectively. We use the ranking-based Spearman correlation rather than Pearson as the relationship between the variables is highly nonlinear (monotonic). The results of Pearson correlation are in Table~\ref{tab:correlation_table2} in the Appendix.}
\label{tab:correlation_table}
\vspace{-0.3cm}
\begin{center}
\begin{scriptsize}
\begin{tabular}{lcccccccccccc}
\toprule
\textbf{Methods}  & Brightness & Contrast & \begin{tabular}{@{}c@{}}Defocus \\ Blur\end{tabular}    &  \begin{tabular}{@{}c@{}}Elastic \\ Transform\end{tabular} & Fog & Frost  & \begin{tabular}{@{}c@{}}Gaussian \\ Blur\end{tabular}   & \begin{tabular}{@{}c@{}}Gaussian \\ Noise\end{tabular}    &  \begin{tabular}{@{}c@{}}Glass \\ Blur\end{tabular}    \\

\midrule
Model ensembles $(+)$ &  $-$0.829 & $-$0.943 &  $-$0.486 &  $-$0.829 &   $-$0.943 &  $-$1.000 &  $-$0.657 &  $-$1.000  &  $-$0.714   \\
MC dropout $(+)$ &  $-$1.000 &  $-$1.000 &   $-$0.600 &   $-$0.943 &  $-$1.000 &   $-$1.000 &   $-$0.829 &  $-$1.000 &  $-$0.486\\
MCInfoNCE  $(-)$ & \textbf{$-$1.000} & \textbf{$-$1.000} & \textbf{$-$0.429} & \textbf{$-$0.943} & \textbf{$-$1.000} & \textbf{$-$1.000} & {$-$0.086} & \textbf{$-$1.000} & {$-$0.714}  \\ 
Ours $(-)$  & \textbf{$-$1.000} & \textbf{$-$1.000} & \textbf{$-$0.429} & \textbf{$-$0.943} & \textbf{$-$1.000} & \textbf{$-$1.000} & \textbf{$-$0.771} & $-$0.600 & \textbf{$-$0.771} \\
\midrule
% \midrule
  &  \begin{tabular}{@{}c@{}}Impulse \\ Noise\end{tabular}  & \begin{tabular}{@{}c@{}}JPEG \\ Comp.\end{tabular}  &  \begin{tabular}{@{}c@{}}Motion \\
 Blur\end{tabular}  & Pixelate & Saturate    & Snow & Spatter & \begin{tabular}{@{}c@{}}Speckle\\ Noise \end{tabular}   & \begin{tabular}{@{}c@{}}Zoom\\ Blur \end{tabular}  \\ 

\midrule
Model ensembles $(+)$ &  $-$1.000 &  $-$1.000 &  $-$0.943 &  $-$1.000 &  $-$0.371 &  $-$0.829 &  $-$0.829 &  $-$1.000 &  $-$0.522   \\ 
MC dropout  $(+)$ &   $-$1.000 &  $-$1.000 &  $-$1.000 &  $-$1.000 &  $-$0.543 &  $-$0.830 &  $-$0.829 &  $-$1.000& $-$0.714        \\
MCInfoNCE  $(-)$ & \textbf{$-$0.943} & \textbf{$-$1.000} & \textbf{$-$0.943} & $-$0.829 & $-$0.371 & $-$0.600 & \textbf{$-$0.829} & \textbf{$-$1.000} & ~~~0.600\\ 
Ours  $(-)$ &  \textbf{$-$0.943} & \textbf{$-$1.000} & \textbf{$-$0.943} & \textbf{$-$0.943} & \textbf{$-$0.714} & \textbf{$-$0.943} & \textbf{$-$0.829} & \textbf{$-$1.000} & \textbf{$-$0.943} \\

\bottomrule
\end{tabular}
\end{scriptsize}
\end{center}
\vskip -0.1in
\end{table*}

\begin{figure*}[t]
    \centering
    \includegraphics[width=1\textwidth]{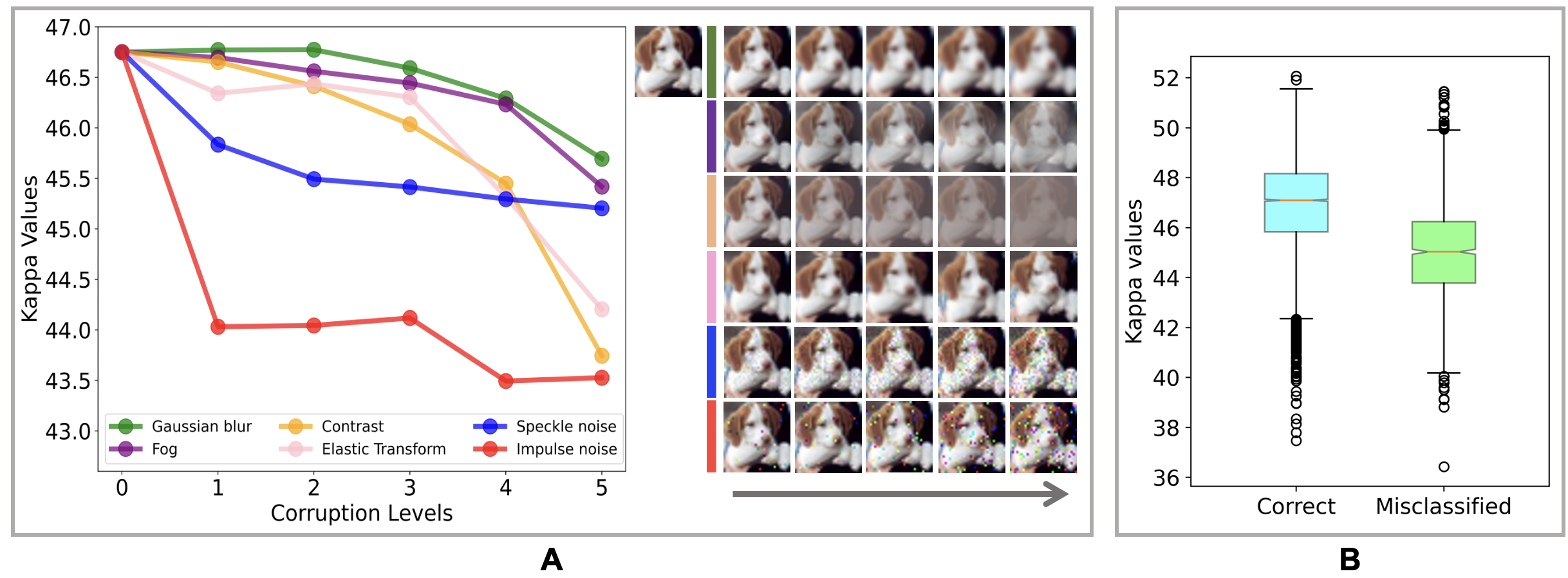}
    \vspace{-0.5cm}
    \caption{\textbf{A.} Decreasing \( \kappa \) implies less concentration and therefore more uncertainty in the representation  (\emph{left}). The associated image corruption is from mild to severe (\emph{right}). \textbf{B.} The two groups of kappa values (i.e., correctly classified and misclassified) from the test set are significantly different.}
    \label{fig:kappa_levels}
\end{figure*}

\subsection{Results}

\textbf{$\mathbf{\kappa}$ captures fine-grained aleatoric uncertainty.}
We validate the efficacy of our framework on CIFAR-10-C \cite{hendrycks2019benchmarking}, particularly in the context of quantifying fine-grained aleatoric uncertainty. The key focus is on how the concentration parameter \( \kappa \) correlates with various levels of data corruption, 
providing a probabilistic interpretation of uncertainty in contrastive learning. 
Table \ref{tab:correlation_table} presents the \emph{Spearman} correlation coefficients between \( \kappa \) and different types of data corruption. Notably, the results reveal a strong correlation in scenarios involving brightness, contrast, and defocus blurring, among others. We observe that model ensembles and MC dropout fail to quantify such fine-grained uncertainty for most of the corruptions. By comparing \emph{MC-InfoNCE} and our method, we observe that \emph{MC-InfoNCE} achieves general good-quality estimation but fails to quantify semantics-related corruptions (such as Gaussian blur and Zoom blur). The formulation of \emph{MC-InfoNCE} enforces the \( \kappa \) for the positive pair to be identical. 
% Since \( \kappa \) is related to the strength of augmentations, naturally this formulation will link \( \kappa \) with the augmentations seen during training, resulting in uninformative \( \kappa \) for unseen image corruptions. In contrast, we enforce the \( \kappa \) to be data-dependent and \emph{explicitly} learn the \( \kappa \) by interacting with their representations. 
Given that \( \kappa \) correlates with the strength of augmentations, this formulation inherently associates \( \kappa \) with augmentations encountered during training, leading to a \( \kappa \) that is uninformative for unseen image corruptions. Our approach ensures \( \kappa \) is dependent on the data and \emph{explicitly} learns \( \kappa \) through interaction with their representations.
Figure~\ref{fig:kappa_levels}(A) offers a visual interpretation of these correlations. It clearly illustrates how the model's uncertainty estimation becomes more pronounced as the data corruption increases, showcasing the sensitivity of our framework in dynamically uncertain environments.

\textbf{$\kappa$ enables failure analysis.} To empirically validate the model's potential in failure analysis, we analyzed the outcome of the CIFAR-10 test set, which includes 10,000 samples. We divided the predictions into two groups: correctly classified (8,554 $\kappa$ values) and misclassified (1,446 $\kappa$ values). The distribution of the two groups is shown in Figure~\ref{fig:kappa_levels}(B). Through bootstrapping (50 iterations, each with 100 randomly sampled observations) and applying the Mann-Whitney U test, we sought to robustly compare $\kappa$ values between the two groups. Our analysis yielded p-values ranging from 
6.42 $\times$ 10$^{-20}$ to 1.15 $\times$ 10$^{-6}$, which strongly suggests a meaningful difference in $\kappa$ values between correctly and incorrectly classified samples, indicating the model's potential in failure detection within practical settings.

% \begin{table}[t]
% \caption{\textbf{AUROC scores for OOD detection.} \emph{ResNet} refers to using a k-NN classifier (k=5) based on ResNet-18 features. The terms \emph{ResNet}+$\kappa$ denotes the enhancement of the classifier through the concatenation of $\kappa$ with the original features, showcasing the impact of leveraging $\kappa$ on the detection capabilities.}
% \vspace{-0.2cm}
% \label{tab:ood}
% \begin{center}
% \scriptsize % Set the font size to footnotesize
% \begin{tabular}{llccccc}
% \toprule
% {In-domain} & OOD & ResNet & $\kappa$ &  ResNet + $\kappa$ \\
% \midrule
% CIFAR-10 &CIFAR-100 & 0.9658 & 0.8162 &\textbf{0.9677}  \\ 
% CIFAR-10 &MNIST & 0.9929   &  0.6783 & \textbf{0.9937}  \\ 
% CIFAR-100 & CIFAR-10 & 0.8653    & 0.6312 & \textbf{0.8794} \\ 
% CIFAR-100 & MNIST & 0.9769    & 0.9390 & \textbf{0.9774}  \\ 
% MNIST&CIFAR-10 & 0.9993  & 0.9979 & \textbf{0.9999}  \\
% MNIST&CIFAR-100 & 0.9998   & 0.9951 & \textbf{1.0000} \\ 
% \bottomrule
% \end{tabular}
% \end{center}
% \end{table}

\begin{table}[t]
    % \centering
    \begin{minipage}[t]{.50\textwidth}
        % \centering
        \label{tab:ood}
        \caption{\textbf{AUROC scores for OOD detection.} $F_\text{res}$ refers to using a k-NN classifier (k=5) based on ResNet-18 features. $F_\text{res}$+$\kappa$ denotes the enhancement through the concatenation of $\kappa$ with the original features.}
        \scriptsize % Set the font size to footnotesize
        \begin{tabular}{llccccc}
        \toprule
        {In-domain} & OOD & $F_\text{res}$ & $\kappa$ &  $F_\text{res}$ + $\kappa$ \\
        \midrule
        CIFAR-10 &CIFAR-100 & 0.9658 & 0.8162 &\textbf{0.9677}  \\ 
        CIFAR-10 &MNIST & 0.9929   &  0.6783 & \textbf{0.9937}  \\ 
        CIFAR-100 & CIFAR-10 & 0.8653    & 0.6312 & \textbf{0.8794} \\ 
        CIFAR-100 & MNIST & 0.9769    & 0.9390 & \textbf{0.9774}  \\ 
        MNIST&CIFAR-10 & 0.9993  & 0.9979 & \textbf{0.9999}  \\
        MNIST&CIFAR-100 & 0.9998   & 0.9951 & \textbf{1.0000} \\ 
        \bottomrule
        \end{tabular}
    \end{minipage}%
    \hfill % This will push the second minipage to the right
    \begin{minipage}[t]{.48\textwidth}
        \begin{minipage}[t]{\textwidth} % Nested top minipage
        \label{tab:other_methods}
            \centering
            \begin{small}
            \caption{\textbf{Extension to other contrastive learning methods}. `Correlation' refers to the average of Spearman correlations in Tab. 1.}
            \end{small}
            \vspace{-0.2cm}
            \scriptsize
            \begin{tabular}{lccccc}
            \toprule
            \textbf{Methods} & SimCLR & SimSiam & BYOL & SwaV \\   
            \midrule
            Correlation & -0.883 & -0.846 & -0.835 & -0.865\\ 
            \bottomrule
            \end{tabular}
        \end{minipage}
        
        \vspace{0.1cm} % Add some space between the tables
        \begin{minipage}[t]{\textwidth} % Nested bottom minipage
            \centering
            \label{tab:dim}
            \begin{tiny}
            \caption{\textbf{The effect of embedding dimensions} with fixed ${\lambda}_{\text{align}}$ and \( \lambda_{\kappa} \). `Correlation' refers to the average of Spearman correlations in Tab. 1.}
            \end{tiny}
            \vspace{-0.2cm}
            \scriptsize
            \begin{tabular}{lccccc}
            \toprule
            \textbf{Dimension} & 64 & 128  & 256 &  384   \\   
            \midrule
            {Correlation} & -0.768 & -0.883 & -0.844 & {-0.901} \\ 
            \bottomrule
            \end{tabular}
        \end{minipage}
    \end{minipage}
\end{table}

% \begin{table}[!htbp]
% \caption{\textbf{The effect of embedding dimensions} with fixed ${\lambda}_{\text{align}}$ and \( \lambda_{\kappa} \). `Correction' refers to the average of 18 Spearman correlations from the types of corruption listed in Table 1.}
% \label{tab:ablation_dim}
% \vskip 0.15in
% \begin{center}
% \begin{small}
% \begin{tabular}{lccccc}
% \toprule
% \textbf{Dimension} & 64 & 128  & 256 &  384   \\   
% \midrule
% {Correlation} & -0.768 & -0.883 & -0.844 & {-0.901} \\ 
% \bottomrule
% \end{tabular}
% \end{small}
% \end{center}
% \vskip -0.1in
% \end{table}

\begin{figure*}[t]
    \centering
    \includegraphics[width=0.94\textwidth]{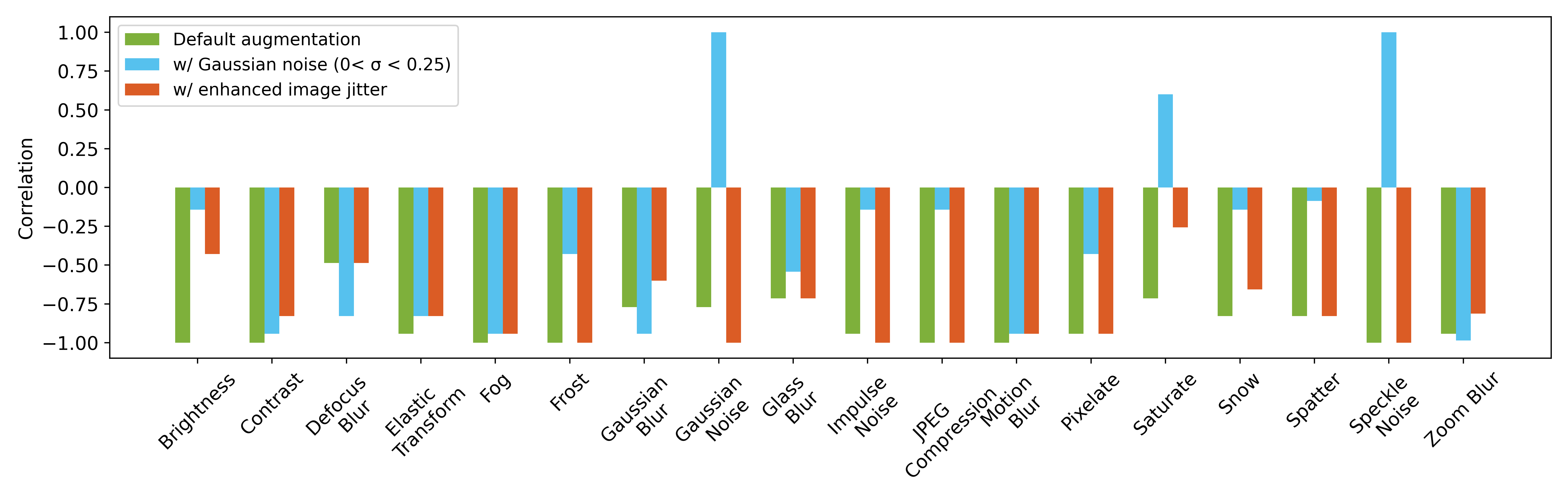}
    \vspace{-0.2cm}
    \caption{\textbf{Additional augmentation degrades the quality of uncertainty estimation for specific types of corruptions}. For instance, introducing Gaussian noise during training causes the correlation with both Gaussian and Speckle noise to shift from negative to positive.}
    \label{fig:kappa_aug}
\end{figure*}

\textbf{$\mathbf{\kappa}$ enhances OOD detection.}
Since $\kappa$ captures inherent characteristics of the data, it may manifest as epistemic uncertainty. The efficacy of $\kappa$ as a self-supervised image feature to \emph{enhance} OOD detection methods is evident from the results presented in Table~2, showcasing consistently superior AUROC values by a simple concatenation with existing features. When compared against \emph{ResNet} feature-based baselines derived from supervised learning approaches as discussed in \cite{ming2022exploit} -- the addition of $\kappa$ consistently enhances performance. This improvement highlights $\kappa$'s capacity to capture aleatoric uncertainty that varied between dataset distributions, thereby validating its utility in strengthening OOD detection methods.

\textbf{$\mathbf{\kappa}$ partially reflects internal augmentations.}
It is known that internal data augmentation during training enables models to learn invariance to those augmentations. Yet, {how the concentration parameter \( \kappa \) reacts to such augmentations}, remains unexplored. We add two types of data augmentations one at a time to test the response of \( \kappa \). Initially, as evidenced by the \greensquare{} bars in Figure \ref{fig:kappa_aug}, the default data augmentations do not weaken the sensitivity of \( \kappa \).   Further introducing Gaussian noise ($\sigma$$<$0.25) into the data augmentation pipeline allows the model to adjust effectively, making \( \kappa \) less sensitive to both Gaussian and speckle noise, as indicated by the \bluesquare{} bars. Furthermore, despite the default augmentation regime, enhancing the image color jittering including brightness (0.3$\rightarrow$0.4), contrast (0.3$\rightarrow$0.4), saturation (0.3$\rightarrow$0.4), and hue (p = 0.2 $\rightarrow$ p = 0.3), \( \kappa \) continues to be reactive to these changes. However, intensifying these augmentations leads to significant shifts in the correlations associated with brightness and similar aspects, highlighted by the \orangesquare{} bars. This suggests the existence of a `saturation point,' beyond which further augmentation fails to meaningfully influence \( \kappa \)'s assessment of uncertainty. Consequently, to preserve \( \kappa \)'s efficacy in uncertainty quantification, our framework advises against the use of overly strong augmentations.

\begin{figure}[t]
    \centering
    \includegraphics[width=0.68\textwidth]{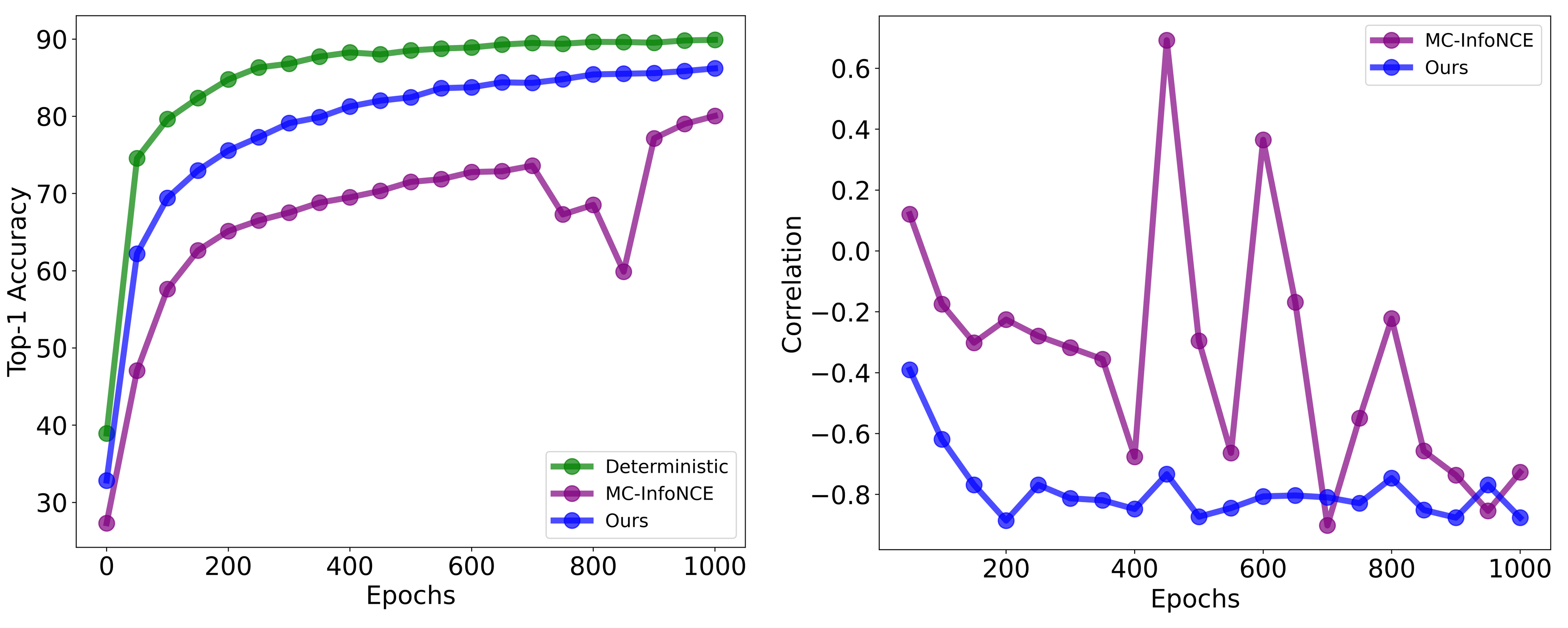}
    \vspace{-0.2cm}
    \caption{\textbf{Left:} Comparison of top-1 classification accuracy on the downstream task over the 1000 training epochs. The deterministic approach represents the original \emph{SimCLR} approach that learns a one-to-one mapping from an image to a representation. \textbf{Right:} Comparison of correlation between \( \kappa \) and levels of data corruption (i.e., uncertainty estimation quality) over the 1000 training epochs.}
    \label{fig:line_epochs}
\end{figure}

\textbf{Integrating uncertainty without losing much discriminativeness.}
Our framework not only models aleatoric uncertainty but also maintains the discriminativeness inherent in contrastive learning models. An analysis depicted on the left panel of Figure \ref{fig:line_epochs} compares the top-1 classification accuracy on the CIFAR-10 test set and the quality of uncertainty estimation across 1000 training epochs. Despite a modest performance decrease (2\%) compared to the deterministic approach, our method exhibits training stability and surpasses the accuracy of the MC sampling-based method \cite{kirchhof2023probabilistic}, demonstrating our model's effectiveness. Furthermore, the right panel of Figure \ref{fig:line_epochs} showcases the consistent performance of our framework in uncertainty estimation. Notably, even in the early stage of training (at the epoch of 200), our model provides high-quality uncertainty estimations. 

\textbf{Adaptability to different methods and dimensions.} We adapt our framework to other established contrastive learning methods such as~\emph{SimSiam} \cite{chen2020simple}, \emph{BYOL} \cite{grill2020bootstrap}, and \emph{SwaV} \cite{caron2020unsupervised}), in a manner of adapting to \emph{SimCLR}. Table 3 demonstrates our framework's versatility, particularly with \emph{SimSiam} and \emph{BYOL}, which train using only positive pairs.  As shown in Table~4, the compatibility of our framework different dimensions of the embedding space further attests to its adaptability. More discussions on the results are in the Appendix.

\section{Discussion}

The findings from this work demonstrate the potential of the concentration parameter \( \kappa \) in uncertainty estimation, failure analysis, and OOD detection. The efficacy of \( \kappa \) in capturing uncertainty, as demonstrated across various datasets and analyses, highlights its applicability in diverse scenarios. However, the scalability of \( \kappa \) in more complex or uncontrolled environments \cite{zhang2023openood,lu2024learning} remains for future work. Further investigation into the integration of \( \kappa \) with other OOD detection methodologies, and validation on new application domains (such as healthcare) could yield further advancements. 

While the results presented herein are promising, this work has its limitations, which nevertheless, open avenues for future research. Pursuing more sophisticated methods to manage \(C_n(\kappa)\) could not only enhance the model's efficacy but also contribute to the broader discourse on probabilistic modeling. Additionally, the exploration of alternative approximations and reparametrization techniques for \( \kappa \) \cite{kato2016robust} could yield deeper insights into the delicate interplay between the probabilistic and deterministic components of contrastive learning. 
In this study, we employ \(\kappa\) primarily for estimating \emph{aleatoric} uncertainty. However, we acknowledge the importance of capturing accurate epistemic uncertainty in the models, to reflect how well the representations are learned in the data-scarcity scenario. We see possible future extensions to incorporate Bayesian inference methods to address this aspect.
Furthermore, the extension of our framework to non-contrastive methods, such as MAE \cite{he2022masked} and larger models (as features are in higher dimensions), and the applications to other types of data, including time series, audio, and text, present exciting challenges. 

% Adapting \( \kappa \) to these modalities may require domain-specific modifications but also promises to broaden the applicability of this approach.

% Such explorations are anticipated to advance our understanding of how models can adapt to varying levels of uncertainty inherent in real-world data.
\textbf{Potential broader impact.}
Our work introduces a new framework for integrating uncertainty estimation into contrastive learning, with significant implications for critical applications such as autonomous driving and medical diagnosis. This advancement supports the development of transparent and accountable AI systems \cite{kim2021machine}, enhancing decision-making quality by providing clearer insights into the model's confidence levels.
Improving uncertainty estimation helps mitigate risks, especially in high-stakes environments, by alerting users when prediction confidence is low. It is crucial to continuously address potential data biases and validate the model against diverse scenarios to ensure reliability. By making AI decisions more interpretable and demonstrating effective ambiguity handling, our approach builds trust among users for widespread adoption.
Future research will focus on integrating this method with other tasks (e.g., classification and segmentation) across various domains, promoting robustness and reliability. %Engaging with stakeholders and adhering to ethical standards during AI development will be vital in preventing misuse and mitigating potential harms.

% The deployment of machine learning systems capable of nuanced uncertainty quantification necessitates careful consideration of their ethical implications and societal impacts. Enhanced decision-making capabilities in critical applications, from autonomous vehicles to medical diagnosis, underscore the importance of developing transparent and accountable systems . Ensuring these systems are interpretable and their decisions understandable to users is paramount in fostering trust and reliability in machine learning. This work not only demonstrates the benefits of our new framework for uncertainty quantification and OOD detection but also sets the stage for a range of future investigations. From refining computational techniques to exploring ethical and societal dimensions, our work creates opportunities to further integrate robustness and reliability into the fabric of machine learning systems.

\bibliographystyle{plain}
\bibliography{ref}

% \printbibliography
%%%%%%%%%%%%%%%%%%%%%%%%%%%%%%%%%%%%%%%%%%%%%%%%%%%%%%%%%%%%

\newpage
\appendix
\section*{Appendix}
\section{Description of corruption types from CIFAR-10-C }

\begin{table}[h]
\centering
\small
\caption{Types of image corruption and their descriptions from CIFAR-10-C \cite{hendrycks2019benchmarking}.}
\begin{tabular}{|l|p{9.5cm}|}
\hline
\textbf{Type} & \textbf{Description} \\
\hline
\textbf{Gaussian Noise} & Often occurs in conditions of poor lighting and adds random fluctuations to pixel values. \\
\hline
\textbf{Shot Noise} & Represents electronic noise emerging due to the inherent discreteness of light, leading to pixel-level variability. \\
\hline
\textbf{Impulse Noise} & Similar to the color version of salt-and-pepper noise, arises from bit errors and manifests as isolated pixel outliers. \\
\hline
\textbf{Defocus Blur} & Occurs when images are not in sharp focus, resulting in a slight blurriness. \\
\hline
\textbf{Frosted Glass Blur} & Resembles the effect seen through frosted glass surfaces, introducing a diffuse and obscured appearance. \\
\hline
\textbf{Motion Blur} & Created by rapid camera movements, causing objects to appear streaked or elongated. \\
\hline
\textbf{Zoom Blur} & Results from quickly moving the camera towards an object, causing a radial blurring effect. \\
\hline
\textbf{Snow} & An obstruction in visual perception, characterized by the presence of white or colored specks in the image. \\
\hline
\textbf{Frost} & Ice crystals on lenses or windows disrupt image clarity, leading to a frosted appearance. \\
\hline
\textbf{Fog} & Cloaks objects in images, simulated using the diamond-square algorithm, resulting in a hazy and obscured view. \\
\hline
\textbf{Brightness} & Affected by variations in daylight intensity, causing overall illumination changes. \\
\hline
\textbf{Contrast} & Depends on lighting conditions and the object's inherent color, leading to alterations in image contrast. \\
\hline
\textbf{Elastic Transformations} & Lead to stretching or contracting of small regions in an image, distorting local features. \\
\hline
\textbf{Pixelation} & A consequence of enlarging a low-resolution image, causing blocky artifacts due to limited pixel information. \\
\hline
\textbf{JPEG Compression} & A lossy method that reduces image size and can introduce artifacts such as blockiness and blurring. \\
\hline
\end{tabular}

\label{table:image_corruption_types}
\end{table}

\newpage
\section{Analysis of the Gradients from $\mathcal{L}_{\text{align}}$}
%%%%%%%%%%%%%%%%%%%%%%%%
% \subsubsection{Analysis of the gradients from $\mathcal{L}_{\text{align}}$}
% According to the chain rule, the gradient of $\mathcal{L}_{\text{align}}$ with respect to $\mathbf{e}_1$ is computed as follows:
\textbf{The gradient of \(\mathcal{L}_{\text{align}}\) w.r.t. \(\boldsymbol{\mu}_1\).}
% We compute the gradient of the exponential alignment loss \(\mathcal{L}_{\text{align}}\) w.r.t. \(\boldsymbol{\mu}_1\) for analysis. %To e need to differentiate the vMF loss function as defined earlier. The vMF loss is given by:
Given \({L}_{\text{a}}\) in Eq. \ref{eqn:vmf}, the gradient of its log w.r.t. \( \boldsymbol{\mu}_1 \) can be obtained by differentiating the loss function w.r.t. \( \boldsymbol{\mu}_1 \).
Its gradient can be expanded as follows:
\begin{flalign}
 \nabla_{\boldsymbol{\mu}_1} \log {L}_{\text{a}} = & \quad \nabla_{\boldsymbol{\mu}_1} \left[ \kappa_1 \cdot \cos(\theta) + \kappa_2 \cdot \cos(\theta) \right]
% \nonumber \\
%  = &\kappa_1 \cdot \nabla_{\boldsymbol{\mu}_1} \cos(\theta) + \kappa_2 \cdot \nabla_{\boldsymbol{\mu}_1} \cos(\theta)
\end{flalign}
Now, \( \cos(\theta) = \boldsymbol{\mu}_1^T \boldsymbol{\mu}_2 \), and its gradient w.r.t. \( \boldsymbol{\mu}_1 \) is \( \boldsymbol{\mu}_2 \). Plug this into the gradient of \(\mathcal{L}_{\text{align}}\), we get:
% \begin{equation}
% \nabla_{\boldsymbol{\mu}_1} \log {L}_{\text{a}} = \kappa_1 \boldsymbol{\mu}_2 + \kappa_2 \boldsymbol{\mu}_2
% \end{equation}
% Finally, plugging this into the gradient of \(\mathcal{L}_{\text{align}}\), we get:
\begin{equation}
\nabla_{\boldsymbol{\mu}_1} \mathcal{L}_{\text{align}} = - \lambda_{\text{align}} \cdot (\kappa_1 \boldsymbol{\mu}_2 + \kappa_2 \boldsymbol{\mu}_2)
\end{equation}
% This gradient represents the direction and magnitude of the change required in \( \boldsymbol{\mu}_1 \) to minimize the exponential alignment loss, taking into account the influence of both \( \boldsymbol{\mu}_2 \) and the concentration parameters \( \kappa_1 \) and \( \kappa_2 \).
This gradient aligns $\boldsymbol{\mu}_1$ towards $\boldsymbol{\mu}_2$, similar to those with existing contrastive losses. More importantly, however, the \textit{strength} of this alignment effect is \textit{controlled} by the estimated concentration parameters $\kappa_1$ and $\kappa_2$ (\textit{i.e.,} the estimated uncertainties) of both $\boldsymbol{\mu}_1$ and $\boldsymbol{\mu}_2$. Smaller $\kappa$'s indicate more uncertainties and lead to looser alignment. Compared with conventional contrastive losses which naively align positive pairs regardless of the severity of corruptions in the input, our $\mathcal{L}_{\text{align}}$ yields a more flexible latent space that is aware of the severity of corruptions in the input. %This gradient ensures the informativeness of estimated uncertainty and mitigates over-confident predictions. 

\textbf{The gradient of \(\mathcal{L}_{\text{align}}\) w.r.t. \(\mathbf{\kappa}_1\).}
Similarly, we can compute the gradient of $\mathcal{L}_{\text{align}}$ w.r.t. $\mathbf{\kappa}_1$ as follows: 
\begin{equation}
\label{eqn:grad_kappa}
\nabla_{\mathbf{\kappa}_1}\mathcal{L}_{\text{align}} = - \mathcal{\lambda}_{\text{align}} \cdot \boldsymbol{\mu}_1^T \boldsymbol{\mu}_2
\end{equation}
Eq.~\ref{eqn:grad_kappa} implies that a closer cosine distance between $\boldsymbol{\mu}_1$ and $\boldsymbol{\mu}_2$ encourages a stronger increase in $\mathbf{\kappa}_1$, indicating reduced uncertainty. The increasing effect on $\mathbf{\kappa}_1$ weakens as the distance between $\boldsymbol{\mu}_1$ and $\boldsymbol{\mu}_2$ grows. Meanwhile, when the angle between $\boldsymbol{\mu}_1$ and $\boldsymbol{\mu}_2$ surpasses $\frac{\pi}{2}$, the gradient encourages a reduction in $\mathbf{\kappa}_1$ instead, hence an increase in predicted uncertainty. Of note, $\mathbf{\kappa}$'s would not grow uninformatively large as they are bounded by the $\ell_2$ regularization (Eq.~\ref{eqn:l2_reg}) at the same time.

% \newpage
\section{Discussion on Results of Different Network Complexity, Embedding Dimensions, and Learning Frameworks}
Table \ref{tab:network} further demonstrates the versatility of our approach across different network architectures, including \emph{ResNet18}, \emph{ResNet34}, and \emph{ResNet50}. Our method consistently achieves strong correlation coefficients, illustrating that the introduction of \( \kappa \) does not compromise the discriminative nature of the embeddings. Instead, it enriches the model's representation by providing a probabilistic dimension that captures uncertainty directly related to the data's intrinsic characteristics.

The compatibility of our framework with established contrastive learning methods, such as~\emph{SimSiam} \cite{chen2020simple}, \emph{BYOL} \cite{grill2020bootstrap}, and \emph{SwaV} \cite{caron2020unsupervised}, further attests to its adaptability.
% This integration not only enhances their robustness and uncertainty quantification capabilities but also maintains their inherent discriminative power, illustrating the flexibility and broad applicability of our approach in managing diverse and complex datasets. 
Table 3 demonstrates our framework's versatility, particularly with \emph{SimSiam} and \emph{BYOL}, which train using only positive pairs. Across these methods, our approach consistently achieves strong correlation coefficients, underscoring the substantial promise of our design. This extension is not merely a testament to the flexibility of our approach but also promises to broaden the applicability of contrastive learning models in handling diverse applications. 

 In Table 4, we investigate the effect of embedding dimensions on  \( {\kappa} \)'s capability to quantify uncertainty. With embedding dimensions set at 64, 128, 256, and 384, our framework demonstrates a nuanced performance variation, indicated by the correlation coefficients -0.768, -0.883, -0.844, and -0.901, respectively. The optimal performance at 128 and 384 dimensions suggests a critical balance between dimensionality and the model's ability to effectively capture uncertainty.
 
\begin{table}[!htbp]
\caption{\textbf{The effect of network complexity} with fixed ${\lambda}_{\text{align}}$, \( \lambda_{\kappa} \), and number of embedding dimension (dim. = 128). `Correction' refers to the average of 18 Spearman correlations from the types of corruption listed in Table 1.}
\label{tab:network}
\vspace{-0.1cm}
\begin{center}
\begin{small}
\begin{tabular}{lccccc}
\toprule
\textbf{Architecture} & ResNet18 & ResNet34  & ResNet50 \\  
\midrule
{Correlation} & -0.908  & -0.876 &  -0.883 \\
% {Accuracy} &  85.640  & 86.610 & 86.720 \\

\bottomrule
\end{tabular}
\end{small}
\end{center}
\vskip -0.1in
\end{table}
%%% 0.848

% \begin{table}[!htbp]
% \caption{\textbf{The effect of embedding dimensions} with fixed ${\lambda}_{\text{align}}$ and \( \lambda_{\kappa} \). `Correction' refers to the average of 18 Spearman correlations from the types of corruption listed in Table 1.}
% \label{tab:ablation_dim}
% \vskip 0.15in
% \begin{center}
% \begin{small}
% \begin{tabular}{lccccc}
% \toprule
% \textbf{Dimension} & 64 & 128  & 256 &  384   \\   
% \midrule
% {Correlation} & -0.768 & -0.883 & -0.844 & {-0.901} \\ 
% \bottomrule
% \end{tabular}
% \end{small}
% \end{center}
% \vskip -0.1in
% \end{table}

% \begin{table}[!htbp]
% \caption{\textbf{Extension to other contrastive learning methods}. `Correction' refers to the average of 18 Spearman correlations from the types of corruption listed in Table 1.}
% \label{tab:architecture}
% \begin{center}
% \begin{small}
% \begin{tabular}{lccccc}
% \toprule
% \textbf{Methods} & SimCLR & SimSiam & BYOL & SwaV \\   
% \midrule
% Correlation & -0.883 & -0.846 & -0.835 & -0.865\\ 
% \bottomrule
% \end{tabular}
% \end{small}
% \end{center}
% % \vskip -0.1in
% \end{table}

% %%%%%%%%%%%%%%%%%%%%%%%

\begin{table*}[!htbp]
\caption{\textbf{Pearson correlation between kappa values and the levels of corruption}. As the severity of corruption increases, the concentration parameter kappa decreases, implying higher uncertainty in the representations. $+$ and $-$ indicate that the correlation are expected to \emph{positive} and \emph{negative}, respectively.}
\label{tab:correlation_table2}
\vspace{-0.4cm}
\begin{center}
\begin{scriptsize}
\begin{tabular}{lcccccccccccc}
\toprule
\textbf{Methods}  & Brightness & Contrast & \begin{tabular}{@{}c@{}}Defocus \\ Blur\end{tabular}    &  \begin{tabular}{@{}c@{}}Elastic \\ Transform\end{tabular} & Fog & Frost  & \begin{tabular}{@{}c@{}}Gaussian \\ Blur\end{tabular}   & \begin{tabular}{@{}c@{}}Gaussian \\ Noise\end{tabular}    &  \begin{tabular}{@{}c@{}}Glass \\ Blur\end{tabular}    \\

\midrule
Model ensembles $(+)$ & $-$0.791 & $-$0.891 & $-$0.687 & $-$0.822 & $-$0.655 & $-$0.986 & $-$0.776 & $-$0.919 & $-$0.685  \\
MC dropout $(+)$ & $-$0.913 & $-$0.882 & $-$0.818 & $-$0.862 & $-$0.920 & $-$0.984 & $-$0.878 & $-$0.916 & $-$0.664 \\
MCInfoNCE $(-)$ & $\textbf{$-$0.915}$ & $\textbf{$-$0.937}$ & $-$0.525 & ${-0.850}$ & $\textbf{$-$0.913}$ & $\textbf{$-$0.969}$ & $-$0.289 & \textbf{$-$0.818} & {$-$0.685} \\ 
Ours $(-)$ & $-$0.893 & $-$0.876 & \textbf{$-$0.802} & \textbf{$-$0.892} & ${-0.901}$ & ${-0.949}$ & \textbf{$-$0.845} & {$-$0.737} & \textbf{$-$0.770} \\ 
\midrule
% \midrule
  &  \begin{tabular}{@{}c@{}}Impulse \\ Noise\end{tabular}  & \begin{tabular}{@{}c@{}}JPEG \\ Comp.\end{tabular}  &  \begin{tabular}{@{}c@{}}Motion \\
 Blur\end{tabular}  & Pixelate & Saturate    & Snow & Spatter & \begin{tabular}{@{}c@{}}Speckle\\ Noise \end{tabular}   & \begin{tabular}{@{}c@{}}Zoom\\ Blur \end{tabular}  \\ 

\midrule
Model ensembles $(+)$ & $-$0.993 & $-$0.918 & $-$0.973 & $-$0.990 & $-$0.350 & $-$0.832 & $-$0.846 & $-$0.976 & $-$0.551 \\ 
MC dropout $(+)$ & $-$0.996 & $-$0.869 & $-$0.987 & $-$0.972 & $-$0.541 & $-$0.835 & $-$0.840 & $-$0.976 & $-$0.825  \\
MCInfoNCE $(-)$ & $-$0.904 & $-$0.869 & $-$0.971 & $-$0.850 & $-$0.462 & $-$0.741 & \textbf{$-$0.892} & \textbf{$-$0.923} & ~0.648 \\ 
Ours $(-)$ & \textbf{$-$0.927} & \textbf{$-$0.959} & \textbf{$-$0.980} & \textbf{$-$0.959} & \textbf{$-$0.563} & \textbf{$-$0.877} & {$-$0.887} & $-$0.881 & \textbf{$-$0.892} \\ 
\bottomrule
\end{tabular}
\end{scriptsize}
\end{center}
% \vskip -0.1in
\end{table*}

\newpage
\section{MC-InfoNCE with the \emph{SimCLR} contrastive loss}
\label{MC-InfoNCE}
\begin{tiny}
\begin{lstlisting}[language=Python]
from torch import nn
import torch
from vmf_sampler import VonMisesFisher
from utils_mc import pairwise_cos_sims, pairwise_l2_dists, log_vmf_norm_const
import torch
import torch.nn as nn

class MCSimCLR(nn.Module):
    def __init__(self, kappa_init=16, n_samples=64, temperature=0.5, device=torch.device('cuda:0')):
        super().__init__()
        self.n_samples = n_samples
        self.kappa = torch.nn.Parameter(torch.ones(1, device=device) * kappa_init, requires_grad=True)
        self.temperature = temperature
        
    def forward(self, mu1, kappa1, mu2, kappa2):
        # Draw samples from the von Mises-Fisher distribution
        samples1 = VonMisesFisher(mu1, kappa1).rsample(torch.Size([self.n_samples])) 
        samples2 = VonMisesFisher(mu2, kappa2).rsample(torch.Size([self.n_samples]))
        # Concatenate positive samples for contrastive loss calculation
        samples = torch.cat([samples1, samples2], dim=1)  # [n_MC, 2 * batch, dim]
        # Compute similarity matrix
        sim_matrix = torch.exp(torch.matmul(samples, samples.transpose(2, 1)) / self.temperature)
        # Create mask to zero-out self-similarities (diagonal elements)
        batch_size = mu1.size(0)
        mask = ~torch.eye(2 * batch_size, device=sim_matrix.device, dtype=bool).repeat(self.n_samples, 1, 1)
        sim_matrix = sim_matrix.masked_select(mask).view(self.n_samples, 2 * batch_size, -1)
        # Similarities for the positive pairs)
        pos_sim = torch.exp(torch.sum(samples1 * samples2, dim=2) / self.temperature)
        pos_sim = torch.cat([pos_sim, pos_sim], dim=1) #Duplicate pos_sim
        loss = -torch.log(pos_sim / sim_matrix.sum(dim=2))
        loss = loss.mean()
        return loss
\end{lstlisting}
\end{tiny}

\newpage
\section{Architecture}
\label{architecture}
\begin{tiny}
\begin{lstlisting}[language=Python]
import torch
import torch.nn as nn
import torch.nn.functional as F
from torchvision.models import resnet50, resnet18, resnet34
class ProbabilisticModel(nn.Module):
    def __init__(self, feature_dim=128):
        super(ProbabilisticModel, self).__init__()

        # Define the layers of the ResNet model
        self.f = []
        for name, module in resnet50().named_children():
            if name == 'conv1':
                module = nn.Conv2d(3, 64, kernel_size=3, stride=1, padding=1, bias=False)
            if not isinstance(module, nn.Linear) and not isinstance(module, nn.MaxPool2d):
                self.f.append(module)
        self.f = nn.Sequential(*self.f)

        # Projection head for feature
        self.g = nn.Sequential(
            nn.Linear(2048, 512, bias=False), 
            nn.BatchNorm1d(512),
            nn.ReLU(inplace=True), 
            nn.Linear(512, feature_dim, bias=True)
        )
        # Additional layer for kappa (concentration parameter)
        self.kappa_head = nn.Sequential(
            nn.Linear(2048, 512, bias=False),
            nn.BatchNorm1d(512),
            nn.ReLU(inplace=True),
            nn.Linear(512, 1, bias=True)  # Outputs kappa for each sample
        )

    def forward(self, x):
        x = self.f(x)
        feature = torch.flatten(x, start_dim=1)
        out = self.g(feature)
        kappa = self.kappa_head(feature)  # Compute kappa for each sample
        # Normalize the feature vector and return it with variance and kappa
        return F.normalize(out, dim=-1), F.softplus(kappa.squeeze(-1)) 
 \end{lstlisting}
 \end{tiny}

\newpage
\section{Training}
\label{training}
\begin{tiny}
\begin{lstlisting}[language=Python]

# train for one epoch to learn the mean vector mu and kappa
def train(net, data_loader, train_optimizer):
    net.train()
    total_loss, total_num, train_bar = 0.0, 0, tqdm(data_loader)
    epsilon = 1e-6  # Small constant for numerical stability
    align_strength = 0.05  # Hyperparameter to regularize the embedding alignment loss
    kappa_reg_strength = 0.005  # Hyperparameter for the regularization strength
    simclr_strength = 1  # Hyperparameter for the strength of SimCLR loss

    for pos_1, pos_2, target in train_bar:
        pos_1, pos_2 = pos_1.to(device), pos_2.to(device)

        mean_1, kappa_1 = net(pos_1)
        mean_2, kappa_2 = net(pos_2)

        # Compute the embedding alignment loss component
        alignment = torch.exp(kappa_1 * F.cosine_similarity(mean_1, mean_2, dim=1)+ \ 
        kappa_2 * F.cosine_similarity(mean_1, mean_2, dim=1))
        align_loss = align_strength * (-torch.log(alignment + epsilon).mean())

        # Compute the regularization loss for kappa (L2 norm)
        kappa_reg_loss = kappa_reg_strength * (torch.mean(kappa_1 ** 2) + \ 
        torch.mean(kappa_2 ** 2))


        # Compute SimCLR contrastive loss
        out = torch.cat([mean_1, mean_2], dim=0)
        sim_matrix = torch.exp(torch.mm(out, out.t().contiguous()) / temperature)
        mask = (torch.ones_like(sim_matrix) - torch.eye(2 * batch_size, \
        device=sim_matrix.device)).bool()
        sim_matrix = sim_matrix.masked_select(mask).view(2 * batch_size, -1)
        pos_sim = torch.exp(torch.sum(mean_1 * mean_2, dim=-1) / temperature)
        pos_sim = torch.cat([pos_sim, pos_sim], dim=0)
        contrastive_loss = simclr_strength * \
        (-torch.log(pos_sim / sim_matrix.sum(dim=-1))).mean()

        # Compute the final loss
        loss = align_loss + contrastive_loss + kappa_reg_loss

        # Backward and optimize
        train_optimizer.zero_grad()
        loss.backward()
        train_optimizer.step()
 \end{lstlisting}
\end{tiny}

\end{document}